# Automatic tongue delineation from MRI images with a convolutional neural network approach

Karyna Isaieva[a]*, Yves Laprie[b], Nicolas Turpault[b], Alexis Houssard[b], Jacques Felblinger[a,c] and Pierre-André Vuissoz[a]

[a]*Université de Lorraine, INSERM, IADI, Nancy, F-54000, France;* [b]*Université de Lorraine, CNRS, Inria, LORIA, Nancy, F-54000, France;* [c]*CIC-IT, INSERM, CHRU de Nancy, Nancy, F-54000, France*

*karyna.isaieva@univ-lorraine.fr

# Automatic tongue delineation from MRI images with a convolutional neural network approach

Tongue contour extraction from real-time magnetic resonance images is a nontrivial task due to the presence of artifacts manifesting in form of blurring or ghostly contours. In this work we present results of automatic tongue delineation achieved by means of U-Net auto-encoder convolutional neural network. We present both intra- and inter-subject validation. We used real-time magnetic resonance images and manually annotated 1-pixel wide contours as inputs. Predicted probability maps were post-processed in order to obtain 1-pixel wide tongue contours. The results are very good and slightly outperform published results on automatic tongue segmentation for intra-subject prediction, and the model performance remains acceptable also for inter-subject prediction.

Keywords: magnetic resonance imaging; convolutional neural network; tongue segmentation; speech articulation; U-Net

## Introduction

Speech production is an eminently dynamic process and the study of articulatory gestures is therefore a central research topic. For this reason, many dynamic data acquisition devices have been developed, including electromagnetic articulography (Kaburagi, Wakamiya, and Honda 2005) and ultrasound. Only X-rays and magnetic resonance provide a global view of the vocal tract.

Unlike X-rays, magnetic resonance imaging does not present any health hazard for the subjects and is therefore an essential tool. More recently, real time MRI (rt-MRI) has appeared and offers high enough acquisition rate to study speech production. Raw images cannot be exploited easily and for this reason it is necessary to extract the articulators' contours from images. A great deal of efforts (Berger and Laprie 1996; Thimm 1999; Jallon and Berthommier 2009) has been put into designing automatic tracking algorithms for X-ray image processing, but the poor quality of those images has never led to acceptable results, and the tracking has often been done by hand when a good precision was necessary. This acceptable solution for X-ray images which are in small numbers is not at all acceptable for rt-MRI images which are available in very large numbers (about 200,000 images for one hour of speech recorded at 50Hz whereas the largest MRI databases have several tens of hours of speech which is our case).

Contrary to X-ray images, there is no superimposition of organs in MRI images, which are tomographic slices of a small thickness. The contouring is therefore easier and computer vision techniques, very often based on snakes methods supplemented by optical flow calculation similar to presented in (Lee et al. 2006; Xu et al. 2016) can be used with the expected advantage of requiring only the manual delineation of the first image. More recent techniques such as active shape models were applied for the delineation of speech articulators (Labrunie et al. 2018) from real-time radial FLASH MR images (Uecker et al. 2010). This approach requires a training database and were compared against other techniques like multiple linear regression or shape particle filtering.

However, there are blurring or ghosting effects in rt-MRI images. Blurring mainly appears because of the relatively large slice thickness (generally 8 mm) resulting in a partial volume effect, i.e. the fact that one part of the slice volume corresponds to flesh and the other to air, especially when the tongue groove is marked. Moreover, displacement of the articulators during the acquisition of each image leads to motion artifacts (Frahm et al. 2014) and the undersampling which speeds up the acquisition causes some specific artifacts.

In the case when the tongue tip is rapidly approaching the teeth to articulate a dental sound, when there is a contact between the tongue and the palate and other comparable situations the artifact manifestation is especially strong and some interpretation is required to choose the threshold intensity separating the tongue tissues with the surrounding air. A deep learning method is therefore a natural choice. Multiple studies based on supervised and unsupervised deep learning dealt with the segmentation of the articulators for MRI and ultrasound images. Air-tissue boundary segmentation of the vocal tract real-time images from publicly available datasets was done with convolutional neural networks (CNN) (Valliappan, Mannem, and Ghosh 2018; Somandepalli, Toutios, and Narayanan 2017). Similar experiments exploiting a CNN approach (Zhu, Styler, and Calloway 2018) or a autoencoding approach (Jaumard-Hakoun et al. 2015) were carried out on ultrasound images with the difficulty that the tongue contour is partially visible. In (Eslami, Neuschaefer-Rube, and Serrurier 2019) the segmentation of good quality static MRI images is achieved with conditional generative adversarial networks.

In our study, we concentrate on the extraction of the tongue contours from real-time MR images and the corresponding learning scheme and overall strategy to easily exploit results as curves and not masks. The question addressed in this work is whether learning can be designed to use 1-pixel wide contours as a segmentation mask and then lead to an acceptable delineation quality with very few spurious contour points. In our case we chose U-Net auto-encoding CNN (Ronneberger, Fischer, and Brox 2015) which turned out to be very efficient for biomedical images.

**Materials and Methods**

We used real-time images from the ArtSpeechMRIfr database (Douros et al. 2019). The subjects are two native French speakers, both males of 35 and 32 years which will be denoted below as S1 and S2, respectively. The images were acquired with a Siemens Prisma-fit 3T scanner (Siemens, Erlangen, Germany). We used the radial RF-spoiled FLASH sequence (Uecker et al. 2010) with TR = 2.02 ms, TE = 1.28 ms, FOV = 19.2×19.2 cm, flip angle = 5 degrees, and the slice thickness is 8 mm. Pixel bandwidth was 1600 Hz/pixel and image resolution 136×136. The acquisition time varied from 34 sec to 90 sec, mostly about 60 sec, so that a subject could have some breaks. At the same time this duration is sufficient to contain several sentences, thus the chosen time intervals represent the compromise between comfort and efficiency. We followed the protocol described in (Niebergall et al. 2013). Images were recorded at a frame rate of 55 frames per second and reconstructed with the algorithm presented in (Uecker et al. 2010).

For our analyses we consider a subset of 600 images of speaker S1. They include 400 consecutive images corresponding to one sentence (originally intended to be used in articulatory copy synthesis (Laprie et al. 2013) experiments) and some manually selected non-similar images of the same speaker. Also we took 100 random images of speaker S2. To determine the ground truth of the tongue contours, all images were delineated manually by 4 investigators and then checked and corrected by the first co-author. The data of S1 was randomly divided into training, validation and testing sets (400, 100 and 100 images respectively), and the images of S2 were used only for testing. Apart from the delineated images which represent a small subset of the database, the algorithm was visually tested on the large part of the ArtSpeechMRIfr database. The segmentation task consisted in classifying the image pixels as contour or non-contour. For this, the contour delineated by hand was transformed into a binary mask, with a 1-pixel wide curve approximating the tongue contour. The pre-processing included image cropping and histogram equalization.

We used the Keras framework (Chollet and others 2015) for the U-Net implementation and a version with skip connections of the U-Net architecture. The initial size of the images was 128×128 and all numbers of filters are divided by 2 compared to (Ronneberger, Fischer, and Brox 2015). We also used zero-padding and binary cross-entropy (Xie and Tu 2015) as a loss function. The model was trained with the Adam optimizer with a minimal learning rate of 1.e-5. Since the class population is very unbalanced the samples should be weighted. The sample weights were adjusted from the validation set and were set at 0.8 and 0.2 for the tongue contour class and non-contour class, respectively. The batch size was 8 and the number of epochs was defined automatically by early stopping with a patience of 10. Since in some cases the model tended to fall into a local minimum where all the points were classified as non-contour, we initialized the model weights obtained by learning with another set of the learning parameters from the same set of images which demonstrated reasonable (but not optimal) prediction result on the validation set. All the parameters (batch size, number of epochs, sample weights, decision threshold) were defined from the validation set only.

      The output of the prediction is a probability map which has to be post-processed. Indeed, contrary to region segmentation where thresholding gives the expected result directly, here thresholding generates chunks of contours separated by gaps. The contours are generally thin, i.e. 1 pixel wide, with some thicker parts and sometimes with small additional spurious clouds of points. These spurious groups of points have to be discarded and contour chunks have to be transformed into a true curve. Therefore, we developed a post-processing algorithm. The first step consists of filtering out outliers. Then, the two contour extremities are found as two adjacent contour points having the greatest angular distance with respect to the contour gravity centre. Then a graph was constructed by connecting the points whose distance was less than the distance between the two extremities minus 1 pixel. Applying Dijkstra's shortest path search algorithm between the two extremities with the quadratic distance as a cost provides the tongue contour. An example of the post-processing is shown on Figure 1.

Unlike the method proposed in (Zhang and Suen 1984), this algorithm deals with gaps and despite its simplicity appeared to give quite precise results. However, these two methods can be combined in future work.

For the same reason, F1 score or Dice coefficient, widely used for evaluation of the segmentation quality, are not applicable for the current problem since even a small difference in a contour position leads to a huge penalization with such an evaluation metrics. To avoid this, we used the Mean Sum of Distance metrics (MSD) (Li, Kambhamettu, and Stone 2005):

$$MSD(U,V) = \frac{1}{n1+n2}\left(\sum_{i=1}^{n1} \min_j |u_i - v_j| + \sum_{i=1}^{n2} \min_j |u_j - v_i|\right),$$

where $u_i \in U$ and $v_i \in V$ are ground truth and predicted curves, $n1$ and $n2$ is a total number of points in corresponding curves and taking minimal distance value refers to the definition of distance between a point and a curve.

To validate our results, we performed a 6-fold cross-validation for speaker S1. We kept the same size for the training, validation and testing sets (i.e. respectively 400, 100 and 100), and all the hyperparameters stayed constant but the number of epochs. The latter was defined automatically from the validation set by early stopping for each fold of the cross-validation independently. The number of epochs chosen for each cross-validation fold is given in Table 1. The model trained from each iteration of 6-fold cross-validation was also tested on the images of S2 separately.

**Results and discussion**

The resulting mean squared distances and corresponding standard deviations between manually delineated and predicted contours for all iterations of 6-fold cross-validation are presented in Table 1. In general, intra-subject prediction proved to be very good since the error is less than 0.7 mm. Examples of the best prediction are shown on Figure

2 (a) and an example video of the prediction on non-labeled data is available (*Example of the Automatic Tongue Delineation for S1*). The biggest errors appear in some cases when there is a contact between the tongue and the hard palate or if the sublingual cavity is small and not clearly visible (Figure 2 (b)). Some minor discrepancies are present due to the difficulty of localizing the exact boundary between tissues and air. From Figure 1 (a) it can be seen that the tongue contours are not sharp and can give rise to several interpretations. This phenomena is primarily explained by the relatively large slice thickness and the presence of partial volume effect as a consequence. Also, some motion artifacts due to the finite acquisition time create some interpretation difficulties.

| # iteration | Epoch number | MSD valid (mm) | MSD test (mm) | MSD S2 (mm) | Mean MSD (mm) |
|---|---|---|---|---|---|
| 1 | 32 | 0.63 ± 0.21 | 0.62 ± 0.17 | 1.29 ± 0.36 | 0.95 ± 0.43 |
| 2 | 35 | 0.64 ± 0.24 | 0.62 ± 0.21 | 1.13 ± 0.32 | 0.88 ± 0.38 |
| 3 | 56 | 0.62 ± 0.20 | 0.60 ± 0.16 | 1.16 ± 0.39 | 0.88 ± 0.41 |
| 4 | 27 | 0.66 ± 0.24 | 0.66 ± 0.25 | 1.40 ± 2.49 | 1.04 ± 1.82 |
| 5 | 54 | 0.64 ± 0.21 | 0.64 ± 0.18 | 1.16 ± 0.29 | 0.91 ± 0.36 |
| 6 | 29 | 0.63 ± 0.17 | 0.64 ± 0.21 | 1.12 ± 0.32 | 0.88 ± 0.35 |

Table 1. Results of 6-fold cross-validation. Number of epochs defined by the early stopping and mean value ± standard deviation for MSD values.

Less accuracy was expected for the inter-subject prediction. The principal source of discrepancy is small misplacements of contours due to the different choice of the threshold intensity (see video (*Example of the Automatic Tongue Delineation for S2*) for an example of prediction on the non-labeled data of S2). Typical examples of the successful prediction are shown on Figure 3 (d) and the best prediction is shown on Figure 3 (a). Sometimes in case of a very tight contact between the soft palate and the tongue, the former was incorrectly classified as a part of the tongue and in some cases epiglottis tissues were predicted as the tongue boundary. Typical errors are presented on Figure 3 (c). One contour of S2 (from 6×100 predictions) was mostly cropped because of the post-processing and thus gave an enormous MSD error (Figure 3 (b)).

The average mean squared distance between manually delineated and predicted contours was 0.92 ± 0.83 mm and 0.90 ± 0.39 if the worst contour is excluded. These results show promising performance for both intra- and inter-speaker prediction. The mean MSD value for intra-subject prediction was 0.63 mm which slightly outperform existing methods of the tongue delineation. In (Labrunie et al. 2018), the best result for the tongue delineation was MSD = 0.68 mm for prediction by modified active shape models. In (Eslami, Neuschaefer-Rube, and Serrurier 2019) precision of MSD 0.8 ± 0.3 mm was reached by means of generative adversarial networks which is more preferable due to the absence of manual contouring. However, it should be noted that in this case high quality static images which do not suffer from motion or reconstruction artifacts were used.

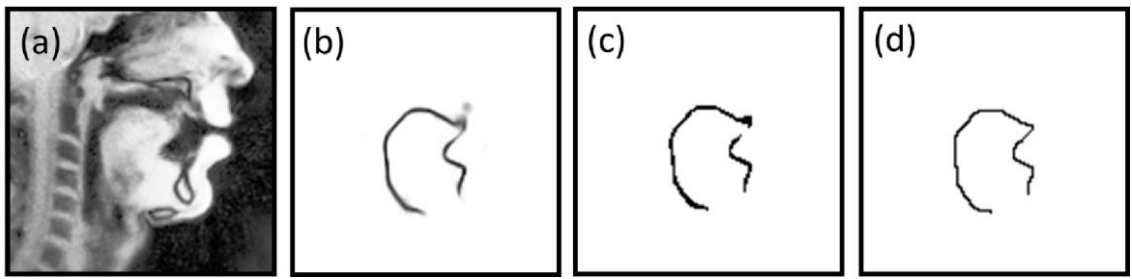

Figure 1. Example of post-processing steps. (a) Initial image. (b) Predicted probability map. (c) Result of thresholding application (decision threshold set to 0.4 for this illustration). (d) Result of the final post-processing.

Taking into account the fact that a limited number of subjects and a rough methodology has been used to select the training images, it is likely that the results of the current work can be improved in the near future. Different choice of the threshold intensity of the tongue for the second subject is probably related to anatomical differences between the speakers. One should understand that due to the relatively large slice thickness a mid-sagittal slice contains not only strictly mid-sagittal contour of the tongue, but also some combination of all para-sagittal contours containing inside the slice. Variability of the tongue shape in left-right direction leads to different level of partial volume effects manifestation and, consequently, to different thicknesses of the intermediate region between the lighter tongue tissues and darker air regions on images. Our further work will engage more speakers with higher anatomical variability. Also, the training data should be chosen more carefully, since the consecutive images do not add large diversity into the training set. Finally, the training set should include more images with a tight contact between the tongue and other articulators to improve the algorithm performance in case of silence and some palatal or velar consonants.

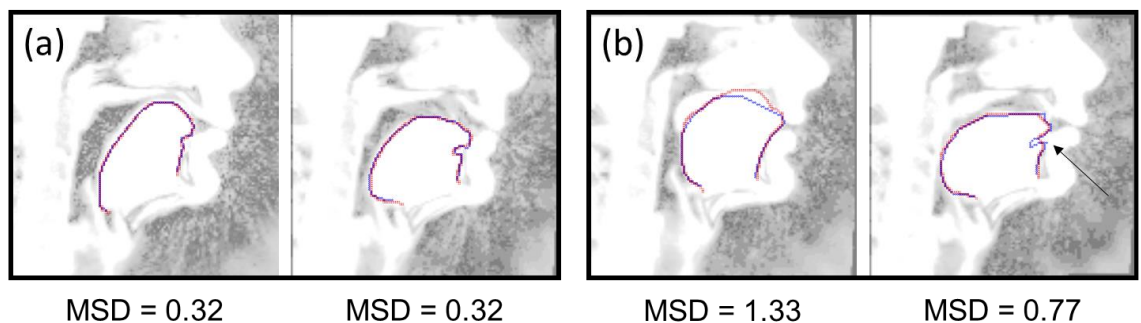

Figure 2. (color online) Examples of the tongue contour prediction in comparison with the manual delineation for speaker S1. Image brightness is changed for better visibility. Predicted contours are denoted by the lighter (red) curves, manually delineated contours are denoted by the darker (blue) curves. (a) Best examples. (b) Typical errors: hard palate is captured - left image, the bend in the region of the front genioglossus muscles is excluded (denoted by the arrow) - the right image.

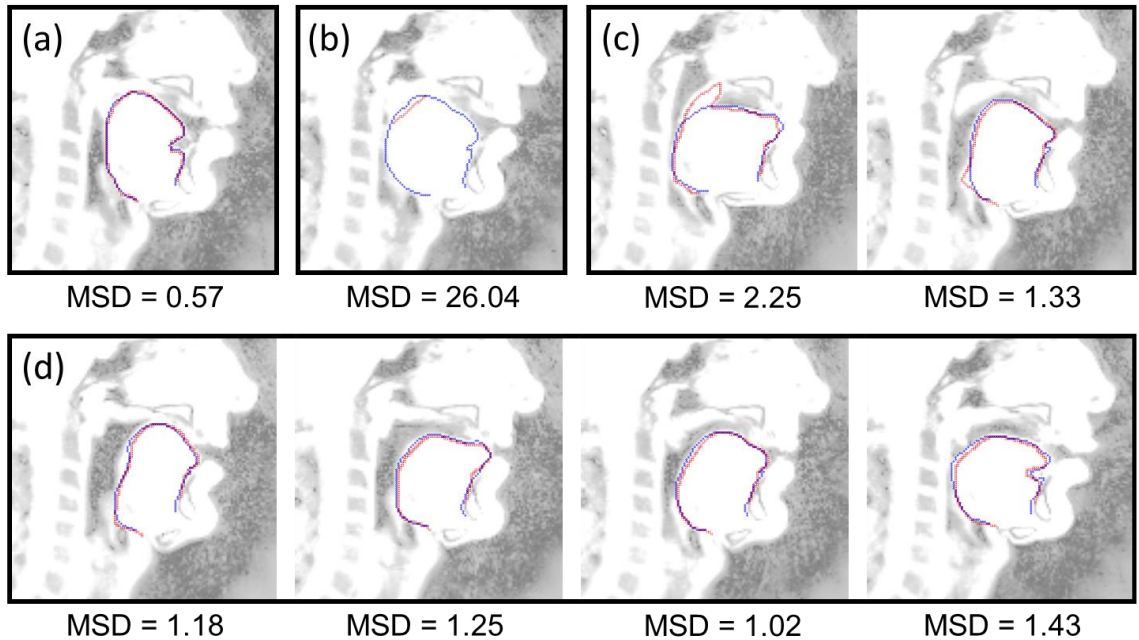

Figure 3. (color online) Examples of the tongue contour prediction in comparison to the manual delineation for the speaker S2. Image brightness is changed for better visibility. Predicted contours are denoted by the lighter (red) curves, manually delineated contours are denoted by the darker (blue) curves. (a) Best example. (b) Worst example (the post-processing failure). (c) Typical cases of the prediction failure: the soft palate is captured-- the left image, the epiglottis is captured - the right image. (d) Typical examples of the algorithm output for the inter-subject prediction.

**Conclusion**

This work shows that a very good delineation accuracy, i.e. less than 0.7mm (for intra-speaker validation), can be achieved for the tongue contour with a rather limited training size if we consider that only 600 images have been used in the training set compared against the 198000 images recorded for this speaker. This approach thus shows promising results and slightly outperforms existing methods despite the presence of multiple image artifacts. This result is all the better since the training set contains 400 consecutive images which were dedicated to carry out articulatory copy synthesis experiments.

Also, we took care to control the learning process in a way to get 1-pixel wide contours and very few spurious points. An additional advantage is that this avoids the use of a skeletization algorithm and only a graph path algorithm is needed to transform the set of points into a curve. The overall strategy is thus simple.

The tests carried out on the second speaker showed that the average accuracy is slightly less (1.21 mm) but still quite acceptable. This means that it is possible to move from one speaker to the other with only a small number of additional training images and probably no further images once the training set will cover a sufficient speaker variability. Further research will focus the other speech articulators and algorithms intended to minimize the set of images that have to be hand labelled.

**Acknowledgment:** Research supported by the project ArtSpeech of ANR (Agence Nationale de la Recherche), France, CPER "IT2MP", "LCHN" and FEDER. We thank Arun A. Joseph, Dirk Voit and Jens Frahm for their help with the data acquisition; Hamza Taybi for his help with the manual contour delineation.

*Example of the Automatic Tongue Delineation for S1.*

*Example of the Automatic Tongue Delineation for S2.*

239.

Zhu, Jian, Will Styler, and Ian C Calloway. 2018. "Automatic Tongue Contour Extraction in Ultrasound Images with Convolutional Neural Networks." *The Journal of the Acoustical Society of America* 143 (3). Acoustical Society of America: 1966.